\documentclass[11pt]{article}

\usepackage[preprint]{acl}

\usepackage{times}
\usepackage{latexsym}
\usepackage{subfig}
\usepackage{multirow} 
\usepackage[T1]{fontenc}
\usepackage{pgf} 

\usepackage[utf8]{inputenc}

\usepackage{microtype}

\usepackage{inconsolata}

\usepackage{graphicx} 
\usepackage{subcaption} 
\usepackage{url}            
\usepackage{booktabs}       
\usepackage{nicefrac}       
\usepackage{amsmath}
\usepackage{amsfonts}
\usepackage{amsthm}
\usepackage{dsfont}
\usepackage{array}   
\usepackage{algorithm}
\usepackage{algpseudocode}
\usepackage{enumitem}

\usepackage{adjustbox}
\usepackage{booktabs}
\usepackage{threeparttable}
\usepackage[dvipsnames,table]{xcolor}
\usepackage{colortbl}
\usepackage{multirow}

\usepackage{tabularx}
\usepackage{natbib}
\usepackage[most]{tcolorbox} 
\definecolor{myblue}{HTML}{499BC0}
\definecolor{myred}{HTML}{F78779}
\definecolor{myblueedge}{RGB}{40,60,130} 
\definecolor{myblueback}{RGB}{235,242,255} 
\definecolor{myblue1}{HTML}{015696}
\definecolor{myblueedge1}{HTML}{0B75B3}
\definecolor{text_red}{HTML}{982B2D}
\definecolor{image_blue}{HTML}{012A61}
\definecolor{positive}{HTML}{008000} 
\definecolor{negative}{HTML}{FF0000}
\definecolor{blue4}{HTML}{335372}
\definecolor{red4}{HTML}{E25659}

\usepackage{hyperref}
\usepackage{todonotes}

\newtcolorbox{keytakeawaybox}{
  enhanced,
  colback=myblueback, 
  colframe=myblueedge, 
  fonttitle=\bfseries\normalsize,
  coltitle=white, 
  title=\textit{Key Takeaway}, 
  attach boxed title to top left={xshift=2mm,yshift=-2mm},
  boxed title style={
    colback=myblue1, 
    rounded corners, 
    boxrule=0pt,
  },
  arc=1mm, 
  boxrule=0.8pt,
  top=3mm,
  bottom=3mm,
  left=3mm,
  right=3mm,
  drop shadow=black!30!white, 
}

\newcommand{\method}{ReAlign}

\title{Generative Giants, Retrieval Weaklings: Why do Multimodal Large Language Models Fail at Multimodal Retrieval?}

\author{
    {\bf Hengyi Feng$^{1,2}$}, 
    {\bf Zeang Sheng$^{2}$}, 
    {\bf Meiyi Qiang$^{2}$}, 
    {\bf Yang Li$^{3}$}, 
    {\bf Wentao Zhang$^{2,4\space \dagger}$} \\
    $^1$ University of Electronic Science and Technology of China \\
    $^2$ Peking University \space 
    $^3$ Tencent Inc \space
    $^4$ Zhongguancun Academy \\
    \textit{hengyi.feng@std.uestc.edu.cn}, \textit{wentao.zhang@pku.edu.cn}
}

\makeatletter
\def\blfootnote{\gdef\@thefnmark{}\@footnotetext}
\makeatother

\begin{document}
\maketitle
\begin{abstract}
\blfootnote{$^\dagger$ Corresponding Author.}
Despite the remarkable success of multimodal large language models (MLLMs) in generative tasks, we observe that they exhibit a counterintuitive deficiency in the zero-shot multimodal retrieval task.
In this work, we investigate the underlying mechanisms that hinder MLLMs from being effective retrievers.
With the help of sparse autoencoders (SAEs), we decompose MLLM output representations into interpretable semantic concepts to probe their intrinsic behavior.
Our analysis reveals that the representation space of MLLMs is overwhelmingly dominated by textual semantics; and the visual semantics essential for multimodal retrieval only constitute a small portion. 
We find that this imbalance is compounded by the heavy focus of MLLMs on bridging image-text modalities, which facilitates generation but homogenizes embeddings and finally diminishes the discriminative power required for multimodal retrieval. 
We further discover that the specific feature components that contribute most to the similarity computations of MLLMs are actually distractors that greatly reduce retrieval performance.
Building on these insights, we propose \method, a test-time adaptation approach that applies a whitening transformation to adjust the geometry of MLLM representation spaces.
Empirical results show that this simple intervention consistently improves zero-shot multimodal retrieval performance across diverse MLLMs without fine-tuning efforts. 
The code is available at \url{https://github.com/Heinz217/mllm-retrieval-analysis}.

\end{abstract}

\section{Introduction}

\begin{figure}[htbp]
    \centering
    \includegraphics[width=\linewidth]{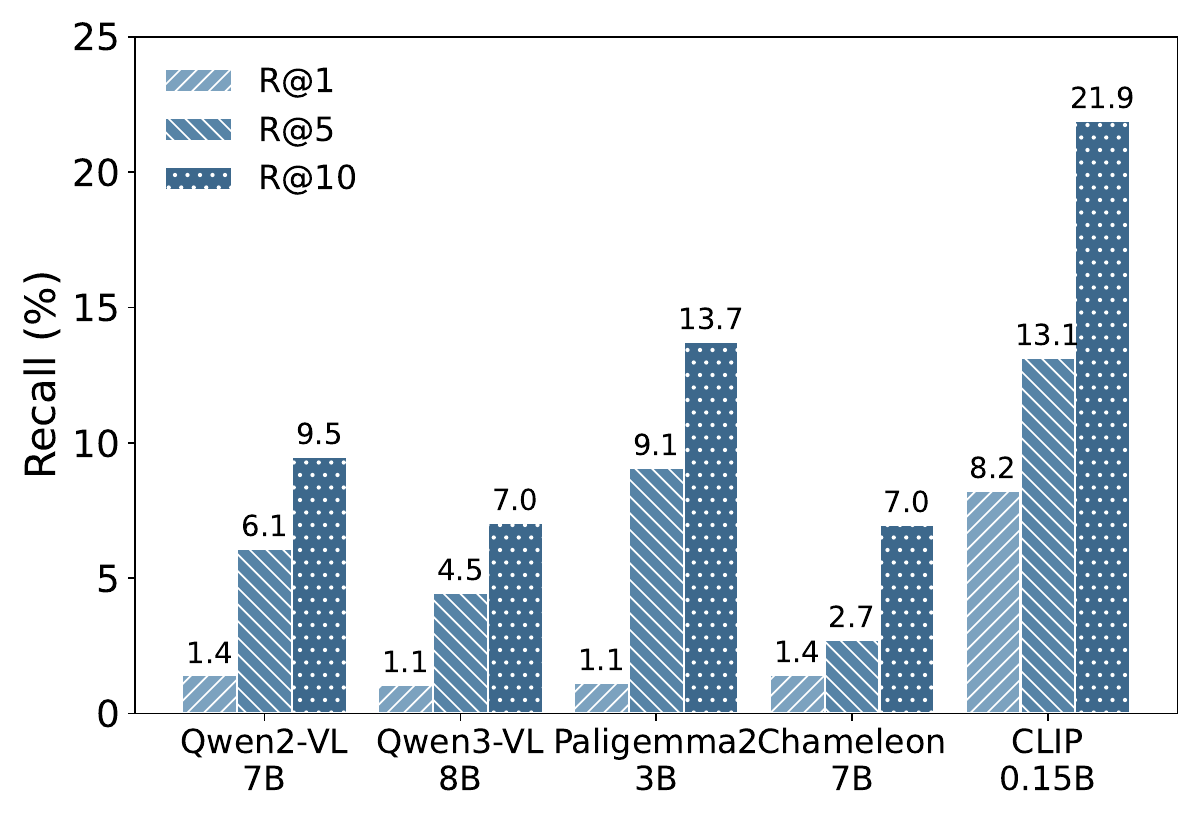}
    \caption{Multimodal retrieval performance of MLLMs and CLIP on the CIRR $((q_i, q_t) \to c_i)$ dataset. The results illustrate the inferior performance of MLLMs.}
    \label{fig:research_question}
    \vspace{-1mm}
\end{figure}

Rapid advances in multimodal large language models (MLLMs)
~\cite{qwen3omni, gemma3, paligemma2, llavanext} 
have revolutionized cross-modality understanding,
establishing state-of-the-art performance in generative tasks such as image captioning, visual question answering (VQA), and complex visual reasoning~\cite{mllmsurvey, bai2024survey}. 
This proficiency mainly stems from MLLM's massive parameter scale~\cite{scalinglaw} that enables the storage of rich parametric knowledge~\cite{knowledge} and the capture of deep semantic dependencies~\cite{wheretolook}. 
Consequently, there is a growing interest in exploring the potential of MLLMs for tasks extending pure generative applications~\cite{VLM2Vec, cai2025lovr}.

As a core task in information retrieval (IR), multimodal retrieval aims to locate relevant content across different data modalities. 
A dominant paradigm within this field is dense retrieval, where queries and candidates (e.g., images and texts) are mapped into a shared high-dimensional space. 
Under this setting, the semantic relationship between modalities is measured by embedding similarity~\cite{denseretrieval2}. 
The models responsible for encoding inputs into vector representations are typically referred to as retrievers.

Intuitively, the extensive parametric memory and the excellent semantic understanding of MLLMs should equip them to act as capable retrievers.
However, in practice, we observe a counterintuitive phenomenon: embeddings directly derived from MLLMs yield poor performance in the zero-shot multimodal retrieval task (Figure.~\ref{fig:research_question}). 
Surprisingly, conventional contrastive vision-language models (e.g., CLIP~\cite{clip}) even significantly outperform MLLMs, although MLLMs have far more parameters than the former.
While MLLMs are not explicitly optimized for representation learning objectives, the magnitude of this performance gap suggests a fundamental limitation when repurposing these generative giants as multimodal retrievers.

Identifying the limitations behind this failure may offer key insights for expanding the use of MLLMs in multimodal retrieval. In this paper, we aim to answer the following research question:

\begin{center}
    \textbf{\textit{What hinders MLLMs from being effective multimodal retrievers?}}
\end{center}

To address this question, we dissect the internal representational mechanisms of MLLMs. 
We employ sparse autoencoders (SAEs)~\cite{sae} as the key analytical tool to disentangle dense representations into linear combinations of interpretable semantic concepts. 
Drawing inspiration from recent work~\cite{colm-sae}, we introduce quantitative metrics to probe the representational space from multiple angles and explicitly link these properties to their impact on the multimodal retrieval performance of MLLMs.
Rigorous empirical experiments that train SAEs on top of billions of activations are conducted on a diverse set of MLLM architectures~\cite{qwen3vl, qwen2vl, Chameleon, paligemma2}.

Through these analyses, we identify three primary factors that limit MLLMs from functioning as effective multimodal retrievers:
\begin{itemize}[leftmargin=*, itemsep=0pt]
    \item 
    We demonstrate that the representational spaces of MLLMs are  
    dominated by the text modality. 
    This inherent bias constrains the capacity to encode visual information essential for retrieval.    
    \item 
    We analyze how MLLMs allocate their representational budget and find a heavy concentration on bridging the modality gap.
    While beneficial for generation, this alignment reduces the distinctiveness of the derived embeddings.
    \item 
    When acting as retrievers, the dominant components on MLLMs's similarity computation turn out to be counterproductive distractors for multimodal retrieval.
\end{itemize}

Guided by these insights, we propose \method, a Test-Time Adaptation intervention designed to rectify the geometry of MLLM feature spaces. 
By applying a whitening transformation, \method\ mitigates representational bottlenecks without requiring additional fine-tuning or labeled data.
Extensive experiments demonstrate that our method unlocks the latent retrieval capabilities of MLLMs, consistently yielding significant retrieval performance improvements across diverse architectures.

In summary, the main contributions and benefits of our work are as follows:
\begin{itemize}[leftmargin=*, itemsep=0pt]
    \item 
    To the best of our knowledge, our work is the first study to analyze MLLM representations and identify intrinsic factors that compromise their performance on multimodal retrieval tasks.
    \item 
    We propose \method, a training-free solution that realigns the embedding spaces of MLLMs, enabling off-the-shelf retrieval performance improvements for MLLMs without finetuning.
\end{itemize}

\section{Preliminary: Zero-shot multimodal retrieval with MLLMs}
\label{sec:preliminary}

In this work, we study the multimodal retrieval task in the zero-shot setting.
Given a query \(q\), the goal is to retrieve a list of candidates \(\{c_1,c_2,\dots ,c_k\}\subset \mathcal{C}\) to maximize ranking metrics such as Recall@1.
The queries and candidates can be text, image, or mixed text–image inputs: \(q\in\{q_t, q_i, (q_t, q_i)\}\); \(c\in\{c_t, c_i, (c_t, c_i)\}\). 
We use the hidden states of multimodal large language models (MLLMs) to extract embeddings for both queries and candidates.

Although MLLMs can produce hidden state representations, they are not originally designed as embedding models.
Previous works~\cite{LamRA, mm-embed} typically use the hidden state of the last token (or special tokens) as the embedding, but such representations work well only when the model has been explicitly optimized for representation learning.
In the zero-shot setting considered here, where no such optimization is allowed, these embeddings are suboptimal.
Instead, we observe that mean pooling over all hidden states yields a consistently better performance (see Appendix~\ref{sec:embedding} for details).
In this work, unless otherwise specified, we adopt mean pooling over the hidden states of the last layer as the embedding for all MLLM-based multimodal retrieval experiments.

\section{What hinders MLLMs from being effective retrievers? A mechanistic analysis with sparse autoencoders}
\label{sec:what_hinders}

To analyze why MLLMs underperform in zero-shot multimodal retrieval scenarios, we perform a mechanistic interpretability analysis using sparse autoencoders (SAEs)~\cite{sparsecoding,ksparseautoencoders}. 

SAEs have gained widespread popularity in recent years due to their remarkable ability to interpret language model activations~\cite{towardsmonosemanticity, jumprelusaes}. 
By reconstructing internal representations with sparsely activated features, SAEs disentangle them into semantic concepts~\cite{sae1, kissane2024interpreting, makelov2024towards}. Given the latent representations \(H \in \mathbb{R}^{n\times d}\), the corresponding sparse codes \(Z \in \mathbb{R}^{n\times c}\) are computed as:
\begin{equation}
    Z = \Phi (HW_{enc}^{\top}+b), \ \ \hat{H}=ZD,
\end{equation}
where \(W_{enc}\in \mathbb{R}^{c\times d}\) is the learned weight matrix, \(D\in \mathbb{R}^{c\times d}\) is the learned dictionary, \(b\in \mathbb{R}^{c}\) is the bias vector, and \(\Phi(\cdot)\) denotes a nonlinear activation function such as ReLU. Specifically, each row of \(D\) can be regarded as a distinct concept vector, capturing an interpretable direction. The reconstruction loss of the SAE is then defined as:
\begin{equation}
    \mathcal{L}_{rec}(H) = \|H - \hat{H}\|^{2} + \alpha \|Z\|_{1},
\end{equation}
where the first term ensures faithful reconstruction of the input representation, and the second term enforces sparsity on the latent code through the \(\ell_1\)-norm, controlled by a parameter \(\alpha\).

In our work, we consider representative MLLMs with different architectures, including Qwen3-VL-8B-Instruct~\cite{qwen3vl}, Qwen2-VL-7B-Instruct~\cite{qwen2vl}
, Chameleon-7B~\cite{Chameleon}, and Paligemma2-3B-Mix-224~\cite{paligemma2}.
For contrastive vision language models (VLMs), we consider CLIP (Clip-ViT-Base-Patch32)~\cite{clip} and SigLIP2 (SigLIP2-Base-Patch16-512)~\cite{siglip2}. In our experiments, we employ Top-K SAEs~\cite{topksae} to learn interpretable concept representations from the activations of the COCO~\cite{mscoco} dataset, using the last-layer hidden states for MLLMs and the last-layer embeddings for contrastive VLMs. More implementation details are provided in Appendix~\ref{appendix:sae}.

\subsection{Evaluation metrics of learned concepts}
\label{sec:metrics}

To evaluate and analyze the learned representations, we introduce four distinct metrics: \textbf{\textit{energy}}, \textbf{\textit{modality score}}, \textbf{\textit{bridge score}}~\cite{colm-sae}, and \textbf{\textit{retrieval attribution score}}. Each measure provides unique insights by focusing on a specific dimension of the representational space.

\paragraph{\textit{Energy.}} Energy represents how frequently and strongly a concept is activated across samples. Specifically, for concept \(i\), it is defined as the expected activation magnitude over all samples:
\begin{equation}
    \label{energy}
    \text{Energy}_i = \mathbb{E}_z \left[ z_i \right].
\end{equation}
Concepts with higher energy are activated more frequently or strongly, capturing more dominant or widely shared patterns in the representation space.

\paragraph{\textit{Modality Score.}} 
The modality score quantifies a concept's bias towards text or image.
For concept \(i\), the modality score is computed as follows:
\begin{equation}
    \label{modality}
    \text{ModalityScore}_i =
    \frac{\mathbb{E}_{z \sim \tau} \left[ z_i \right]}
    {\mathbb{E}_{z \sim \iota} \left[ z_i \right] + \mathbb{E}_{z \sim \tau} \left[ z_i \right]},
\end{equation}
where \(\iota\) and \(\tau\) denote the image and text modalities, respectively. 
For a given concept, higher scores indicate text dominance, whereas lower scores indicate image dominance.
Concepts with balanced scores act as shared cross-modality features.

\paragraph{\textit{Bridge Score.}} 
The bridge score \(\mathbf{B} \in \mathbb{R}^{c \times c}\) quantifies the extent to which concepts serve as connectors between the image and the text modalities.
It is defined as: 
\begin{equation}
    \label{bridge}
    \mathbf{B} =
    \mathbb{E}_{(z_{\iota}, z_{\tau})\sim\gamma}[z_{\iota}^\top z_{\tau}]
    \odot (D D^\top),
\end{equation}
where \({(z_{\iota}, z_{\tau})}{\sim \gamma}\) denotes the pair of sparse codes obtained from a matching image-text pair, and \(\odot\) denotes the Hadamard product. 
Higher bridge scores indicate concepts that function as semantic connectors across the two modalities.

\paragraph{\textit{Retrieval Attribution Score.}} The retrieval attribution score \(\mathbf{A} \in \mathbb{R}^{c}\) measures the contribution of concepts to the overall cross-modality similarity between image and text representations. 
The contribution of each concept is decomposed through reconstruction interactions:
\begin{equation}
    \label{attribution}
    \mathbf{A}=
    \mathbb{E}_{(z_{\iota}, z_{\tau}) \sim \gamma}
    \left[z_{\iota} \odot (M z_{\tau})+z_{\tau} \odot (M z_{\iota})\right],
\end{equation}
where \(M = DD^{\top}\). 
The score is tailored for multimodal retrieval settings and measures each concept's impact on cross-modality matching, with higher scores indicating greater influence on the final similarity between queries and candidates.

\begin{figure*}[h]
    \centering
    \includegraphics[width=\textwidth]{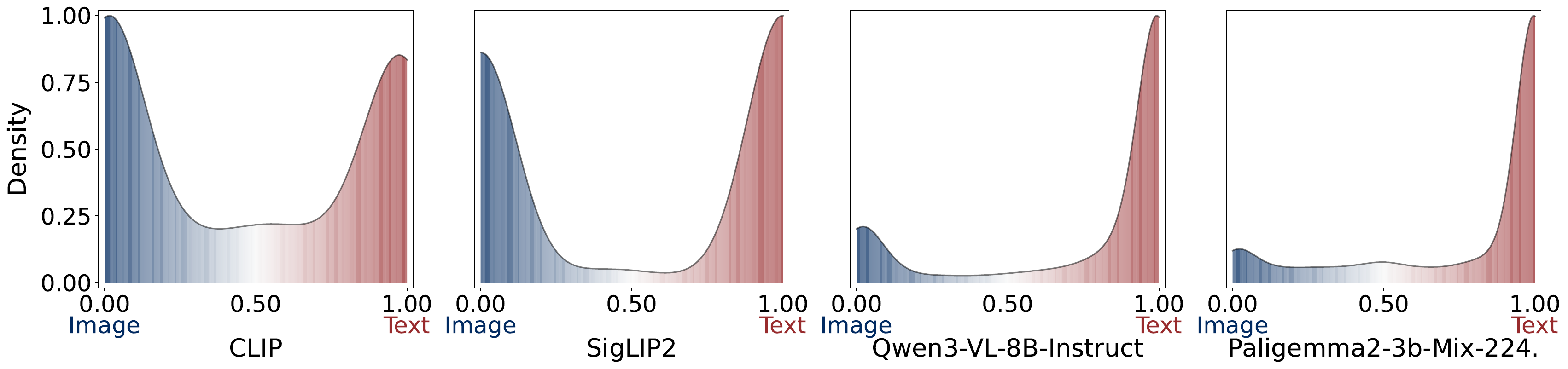}
    \caption{Distribution of \textbf{\textit{modality scores}} for learned concepts by \textbf{(a) CLIP}, \textbf{(b) SigLIP2}, \textbf{(c) Qwen3-VL-8B-Instruct}, and \textbf{(d) Paligemma2-3B-Mix-224}. The Modality Score quantifies the bias of each concept towards the image modality (\textcolor{image_blue}{blue region}) or the text modality (\textcolor{text_red}{red region}). The distributions are visualized using the Kernel Density Estimation (KDE) method~\cite{KDE} based on concept activation statistics.}
    \label{fig:modality}
\end{figure*}

\subsection{Observation: Textual information dominates representations in MLLMs}
\label{gap}

\begin{keytakeawaybox}
\textit{The strong text bias in MLLMs’ representational space limits the encoding of visual information and undermines multimodal retrieval effectiveness.}
\end{keytakeawaybox}

We first analyze the distribution of the \textbf{\textit{modality score}} (Eq.~\ref{modality}).
In all models that we analyze, as shown in Figure~\ref{fig:modality}, our first observation is that the majority of the learned concepts are single-modality. Furthermore, the concepts learned by MLLMs exhibit a strong bias towards the text modality. The distribution of modality scores reveals that a large proportion of concepts are text-specific,
indicating that the representations generated by MLLMs are primarily driven by linguistic information rather than visual information. 

However, unlike MLLMs, the distribution of image- and text-specific concepts is more balanced in CLIP and SigLIP2. Moreover, they exhibit a larger fraction of intermediate concepts, which encode information from both modalities, serving as shared semantic anchors that better connect the visual and textual representations.

Previous studies have discussed the existence of modality bias~\cite{zheng2025mllmsdeeplyaffectedmodality} and modality gap~\cite{mindthegap,eslami2025mitigate} in multimodal models. 
In the context of multimodal retrieval, we argue that maintaining a balanced modality representation is particularly important.
When a model’s representations are heavily biased towards one modality, its embeddings fail to adequately represent samples from the other, leading to degraded retrieval performance under cross-modality or mixed-modality retrieval scenarios. 
Meanwhile, contrastive VLMs, with more balanced and interconnected concept spaces, are more capable of producing modality-agnostic embeddings, achieving better performance in zero-shot multimodal retrieval settings than MLLMs.

\subsection{Observation: MLLMs concentrate most of their representational efforts on bridging image-text modalities}

\begin{keytakeawaybox}
\textit{MLLMs overly align visual information into the text space, producing embeddings that are less distinctive across samples and impairing multimodal retrieval performance.}
\end{keytakeawaybox}

\begin{figure}[t]
    \centering
    \includegraphics[width=\columnwidth]{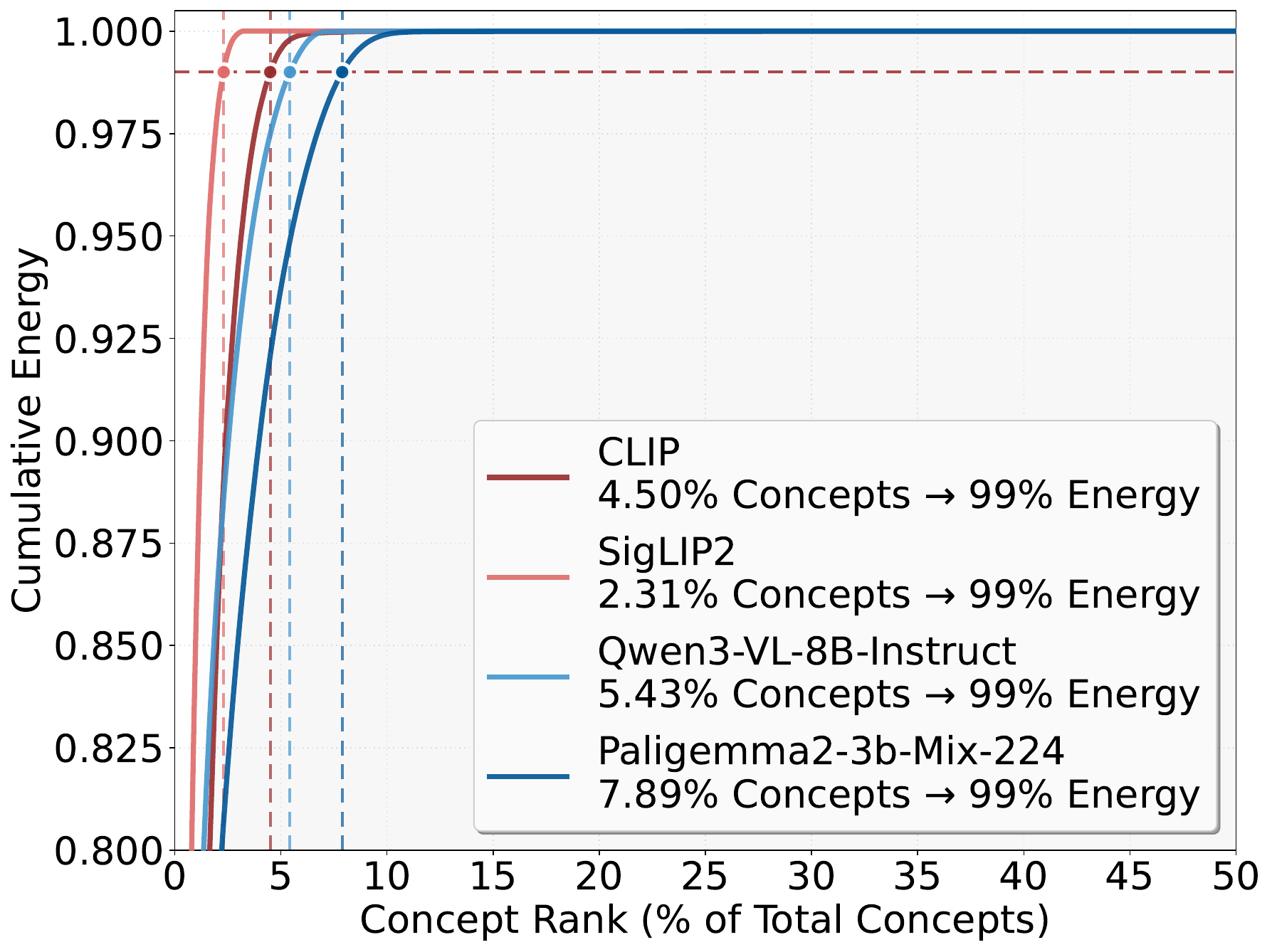}
    \caption{Cumulative \textbf{\textit{energy}} distribution across concept ranks for different multimodal models. The curves show the percentage of total energy captured as concepts are ranked by their individual energy values.}
    \label{fig:energy}
\end{figure}

\begin{table*}[h]
    \centering
    \resizebox{0.9\textwidth}{!}{
    \begin{tabular}{llcccc} 
    \toprule
    \textbf{Type} & \textbf{Model} & \textbf{\(J(\mathcal{S}_E, \mathcal{S}_A)\)} & \textbf{\(J(\mathcal{S}_E, \mathcal{S}_B)\)} & \textbf{\(J(\mathcal{S}_A, \mathcal{S}_B)\)} & \textbf{Triple Overlap} \\
    \midrule
    \multirow{2}{*}{Contrastive VLMs} 
    & CLIP & 0.5674 & 0.6831 & 0.6126 & 0.4796 \\
    & SigLIP2 & 0.3948 & 0.5775 & 0.3783 & 0.2915 \\
    \midrule
    \multirow{4}{*}{MLLMs} 
    & Qwen2-VL-7B-Instruct & 0.6494 & 0.7957 & 0.7146 & 0.5989 \\
    & Qwen3-VL-8B-Instruct & 0.6610 & 0.8547 & 0.6700 & 0.5486 \\
    & Paligemma2-3B-Mix-224 & 0.7655 & 0.8085 & 0.7124 & 0.6541 \\
    & Chameleon-7B & 0.6850 & 0.8591 & 0.7099 & 0.5953 \\ 
    \bottomrule
    \end{tabular}
    }
    \caption{Jaccard similarity between top 1\% concept sets across three different evaluation metrics. \(\mathcal{S}_E\): top \textbf{\textit{energy}} set, \(\mathcal{S}_A\): top \textbf{\textit{retrieval attribution score}} set, \(\mathcal{S}_B\): top \textbf{\textit{bridge score}} set. 
    }
    \label{tab:jaccard}
    \vspace{-2mm}
\end{table*}

When we examine the \textit{\textbf{energy}} (Eq.~\ref{energy}) distribution of the learned concepts, it is observed that for all models, the distribution exhibits a pronounced long-tail pattern (Figure~\ref{fig:energy}). This reveals that during the reconstruction process, only a small subset of concepts is repeatedly activated across different samples. 
These concepts constitute the main concentrations of the model’s representational space.
These concepts therefore play a critical role in shaping the overall embedding structure.

To further explore where this representational energy is allocated, we analyze two additional metrics, \textbf{\textit{bridge score}} (Eq.~\ref{bridge}) and \textit{\textbf{retrieval attribution score}} (Eq.~\ref{attribution}), for a combined assessment.
Our goal is to uncover how multimodal models, particularly MLLMs, distribute their energy across concepts and how such allocation patterns affect their retrieval performance. 
Specifically, for each concept, we compute its energy, bridge score, and retrieval attribution score, and then extract the top \(1\%\) concepts for each metric. 
We measure the Jaccard similarity between these top sets to quantify the degree of overlap, where a higher overlap demonstrates that the same set of concepts dominates multiple representational dimensions.

Formally, for two concept sets \(\mathcal{A}\) and \(\mathcal{B}\), the Jaccard similarity is computed as:
\begin{equation}
    \label{jaccard}
    J(\mathcal{A}, \mathcal{B}) = \frac{|\mathcal{A} \cap \mathcal{B}|}{|\mathcal{A} \cup \mathcal{B}|}.
\end{equation}
As shown in Table~\ref{tab:jaccard}, both MLLMs and contrastive VLMs 
exhibit significant overlaps between the concepts of high-energy, high-bridge, and high-retrieval-attribution score sets. 
However, the overlap ratios are consistently higher in MLLMs.
Notably, the intersection between high-energy and high-bridge concepts is particularly strong.
This suggests that MLLMs devote a large portion of their representational energy to bridging modalities, attempting to harmonize image and text representations within a unified latent space. 

At first glance, this behavior might appear beneficial for multimodal integration and even for retrieval. 
However, when combined with the findings discussed in Section~\ref{gap}, a clearer picture emerges: MLLMs primarily align multimodal information by projecting visual information towards the text space, rather than establishing a balanced semantic fusion. This could lead to an inferior ability of MLLMs to encode visual information.

To validate this, we conduct an additional experiment.
Before mean pooling, we mask out the hidden states corresponding to image tokens and then recompute the final embeddings.

\begin{figure}[htbp]
    \centering
    \includegraphics[width=\linewidth]{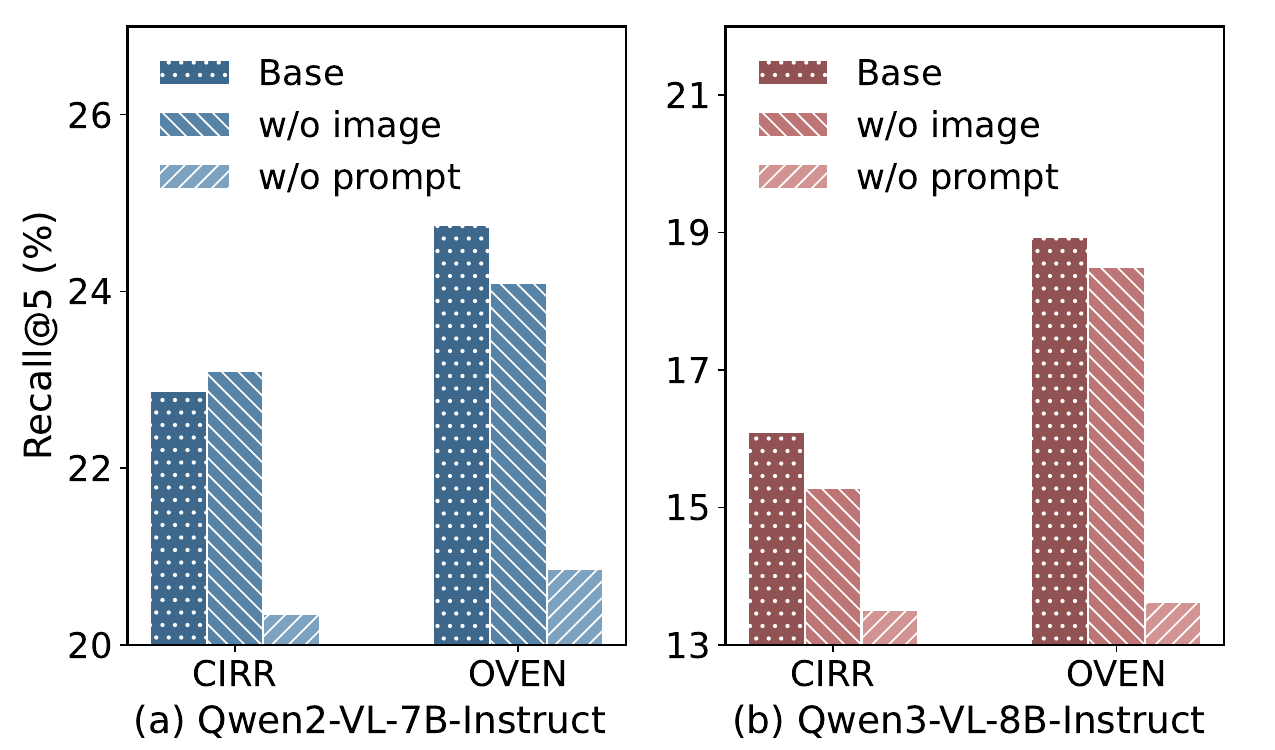}
    \caption{Retrieval performance on the subset (3k queries) of CIRR $((q_i, q_t) \to c_t)$ and OVEN \(((q_i, q_t) \to (q_i, q_t))\). "Base" uses the full input; "w/o image" and "w/o prompt" denote the removal of image tokens and prompt tokens, respectively.}
    \label{fig:token_effect}
    \vspace{-1mm}
\end{figure}

As shown in Figure~\ref{fig:token_effect}, we find that removing image tokens results in only marginal changes in retrieval performance, while masking the user prompt region leads to a sharp performance decline. 

From a retrieval task perspective, effective retrieval performance largely depends on the distinctiveness of embeddings between samples. 
When we examine the overlap between high-retrieval-attribution score concepts (those most relevant to retrieval) and the high-energy / high-bridge score concepts, we observe a significant intersection among concepts learned from MLLMs. 
This suggests that the same dominant concepts not only consume most of the representational energy but also contribute most to multimodal similarity computation for retrieval.
Consequently, the embeddings of different samples become concentrated around similar text-driven features with limited discriminability.

Meanwhile, contrastive VLMs display weaker overlap among these metrics. The high-energy, high-bridge score, and high-retrieval-attribution score concepts are more loosely coupled, allowing them to provide representations with greater discriminability. 
This ultimately contributes to their superior performance in multimodal retrieval. 

\subsection{Observation: Dominant components in similarity computation are counterproductive for retrieval}

\begin{keytakeawaybox}
\textit{When MLLMs act as retrievers, the components exerting the greatest influence on similarity scores are in fact counterproductive distractors for multimodal retrieval.}
\end{keytakeawaybox}

\begin{table}[h]
    \centering
    \resizebox{\columnwidth}{!}{
    \begin{tabular}{lccccc}
        \toprule
        \textbf{Model} & \textbf{Setting} & \textbf{R@1} & \textbf{R@5} & \textbf{R@10} \\
        \midrule
        \multirow{2}{*}{Qwen2-VL-7B} 
        & Base   & 5.60 & 15.44 & 22.66 \\
        & Removal & 13.16 & 29.52 & 39.76 \\
        \midrule
        \multirow{2}{*}{Qwen3-VL-8B} 
        & Base   & 4.71 & 13.86 & 21.34 \\
        & Removal & 17.04 & 37.44 & 48.22 \\
        \midrule
        \multirow{2}{*}{PaliGemma2-3B} 
        & Base   & 5.94 & 14.22 & 19.25 \\
        & Removal & 23.18 & 46.84 & 58.64 \\
        \midrule
        \multirow{2}{*}{Chameleon-7B} 
        & Base   & 1.79 & 7.58 & 10.12 \\
        & Removal & 9.02 & 24.12 & 31.34 \\
        \bottomrule
    \end{tabular}
    }
    \caption{
    Retrieval performance on the MSCOCO $(q_i \to c_t)$ dataset before (Base) and after removing the subspace spanned by concepts with high retrieval attribution scores (Removal), evaluated on Qwen2-VL-7B (\textbf{Qwen2-VL-7B-Instruct}), Qwen3-VL-8B (\textbf{Qwen3-VL-8B-Instruct}), PaliGemma2-3B (\textbf{Paligemma2-3B-Mix-224}), and \textbf{Chameleon-7B}.
    }
    \label{tab:high-attr}
\end{table}

A core objective of retrieval is to compute a reliable similarity metric between query and candidate embeddings. Given the suboptimal performance of MLLMs in multimodal retrieval and the analysis in the previous section, a critical question arises:  
\textit{Do the representation components that exert the greatest influence on the similarity calculation genuinely contribute to retrieval performance?}

To investigate this, we propose an ablation strategy that removes the components corresponding to concepts that exhibit high \textbf{\textit{retrieval attribution scores}} (Eq.~\ref{attribution}) from the representational space.

Concretely, let \(D_R\subseteq D\) denote the sub-dictionary formed by the concept atoms whose retrieval attribution scores fall in the top \(1\%\) (\(m\) concepts). To characterize the subspace \(\mathcal{S}\) spanned by these concepts with high influence, we perform Singular Value Decomposition (SVD) on \(D_R\):
\begin{equation}
    \label{svd}
    D_R = U \Sigma V^\top,
\end{equation}
where \(V \in \mathbb{R}^{d\times m}\) contains the right singular vectors
that form an orthonormal basis for dominant directions in the embedding space.
We select the top \(r\) right dominant singular vectors \(V_r \in \mathbb{R}^{d\times r}\) to construct the basis for the subspace \(\mathcal{S}\).

Given the embedding \(h\in \mathbb{R}^d\) 
, we project it onto \(\mathcal{S}\) to isolate the components associated with \(D_R\):
\begin{equation}
    \label{projection}
    \Pi_{\mathcal{S}}(h)= V_{r} V_{r}^{\top} h.
\end{equation}
We then compute the representation of removing the effect of \(D_R\) by subtracting this projection:
\begin{equation}
    \tilde{h} = h - \Pi_{\mathcal{S}}(h).   
\end{equation}
The resulting residual embedding \(\tilde{h}\) is normalized and utilized directly for retrieval.

The retrieval performance using the transformed embeddings is reported in Table~\ref{tab:high-attr}, and the results reveal that removing the concepts with the highest retrieval attribution scores yields a significant improvement in retrieval accuracy.

This suggests that the components dominating the similarity calculation in  MLLM-derived embeddings are, in effect, counterproductive for retrieval tasks. 
While these concepts possess 
heavy influence on the dot product magnitude, they appear to act as "distractors" that inflate similarity scores regardless of semantic relevance, implying that in the raw representational spaces of MLLMs, the semantically discriminative features necessary for precise retrieval are severely overshadowed.

\section{\method: An efficient test-time solution for MLLM multimodal retrieval}
\label{sec:method}

\subsection{Methodology}

Despite the vast parametric knowledge embedded within MLLMs, our empirical observations reveal that their native representational space is suboptimal for multimodal retrieval, reflected as an imbalance in the distribution of semantic components within the embedding space.
Under this analysis, a natural progression is seeking a transformation of the MLLM's output representation space that rectifies the feature distribution to enhance the semantic distinctiveness of query and candidate embeddings. 

To this end, we propose \method, a training-free \textit{Test-Time Adaptation (TTA)} framework. \method\ operates as a lightweight post-processing module that leverages a whitening strategy to realign the geometry of the feature space on-the-fly. 
This approach mitigates representational bottlenecks without requiring computationally expensive fine-tuning or access to labeled data.

Given the mean-pooled representations \(\bar{h}_i \in \mathbb{R}^d\) derived from the last layer of MLLMs (as described in Section~\ref{sec:preliminary}), \method\ maps them into an isotropic space via a Zero-phase Component Analysis (ZCA) whitening strategy~\cite{ZCA}. 
To ensure numerical stability in high-dimensional settings, we employ a shrinkage estimator for the covariance matrix calculation:
\begin{equation}
    \hat{\Sigma} = (1 - \beta) \Sigma + \beta \frac{\text{Tr}(\Sigma)}{d} I,
\end{equation}
where \(\Sigma\) is the empirical covariance of the centered features, \(\beta \in [0, 1]\) is the shrinkage coefficient, and \(I\) denotes the identity matrix. This regularization balances the empirical covariance with a spherical prior. Here, let \(\hat{\Sigma} = U \Lambda U^\top\) be the eigendecomposition. The final whitened and normalized representation \(e_i\) is computed as:
\begin{equation}
    e_i = \frac{(\bar{h}_i - \mu)^\top U (\Lambda + \epsilon I)^{-\frac{1}{2}} U^\top }{\|(\bar{h}_i - \mu)^\top U (\Lambda + \epsilon I)^{-\frac{1}{2}} U^\top \|_2},
\end{equation}
where \(\mu\) is the mean vector, \(\epsilon\) (e.g., \(10^{-5}\)) is a small constant for numerical stability, and \(\|\cdot\|_2\) denotes \(\ell_2\)-norm normalization. 
This transformation effectively aligns the embeddings to a spherical distribution, enhancing the semantic distinctiveness.

Specifically, we adopt an asymmetric estimation strategy to accommodate the distinct data flows of candidates and queries. 
For the candidate set, we compute the global statistics directly over the entire set to capture its holistic geometry. 
In contrast, we introduce a support set (sampled from the training set) to calculate the statistics for queries since queries arrive in mini-batches where local estimation suffers from high variance.

\begin{table*}[h]
    \centering
    \definecolor{blue4}{HTML}{335372}
\definecolor{image_blu}{HTML}{012A61}
\definecolor{text_red}{HTML}{982B2D}

\resizebox{\textwidth}{!}{%
\begin{tabular}{llrrrrrrrr} 
\toprule
&  & \multicolumn{2}{c}{\textbf{Qwen2-VL-7B}} 
& \multicolumn{2}{c}{\textbf{Qwen3-VL-8B}} 
& \multicolumn{2}{c}{\textbf{Paligemma2-3B}} 
& \multicolumn{2}{c}{\textbf{Chameleon-7B}} 
\\
\cmidrule(lr){3-4}
\cmidrule(lr){5-6}
\cmidrule(lr){7-8}
\cmidrule(lr){9-10}

\textbf{Task} & \textbf{Dataset}
& \multicolumn{1}{c}{Base} & \multicolumn{1}{c}{ReAlign}
& \multicolumn{1}{c}{Base} & \multicolumn{1}{c}{ReAlign}
& \multicolumn{1}{c}{Base} & \multicolumn{1}{c}{ReAlign}
& \multicolumn{1}{c}{Base} & \multicolumn{1}{c}{ReAlign} 
\\

\midrule
\multirow{3}{*}{1. $q_t \to c_i$}
& VisualNews   
& 2.45 & 15.46 \textcolor{text_red}{\scriptsize{+13.01}} 
& 2.16 & 14.89 \textcolor{text_red}{\scriptsize{+12.73}}
& 3.97 & 12.01 \textcolor{text_red}{\scriptsize{+8.04}}
& 1.68 & 7.50 \textcolor{text_red}{\scriptsize{+5.82}}
\\
& MSCOCO       
& 18.57 & 59.80 \textcolor{text_red}{\scriptsize{+41.23}} 
& 16.39 & 57.63 \textcolor{text_red}{\scriptsize{+41.24}} 
& 26.63 & 66.38 \textcolor{text_red}{\scriptsize{+39.75}} 
& 9.74 & 48.50 \textcolor{text_red}{\scriptsize{+38.76}} 
\\
& Fashion200K  
& 6.98 & 9.37 \textcolor{text_red}{\scriptsize{+2.39}} 
& 7.25 & 8.99 \textcolor{text_red}{\scriptsize{+1.74}} 
& 1.37 & 5.06 \textcolor{text_red}{\scriptsize{+3.69}} 
& 1.87 & 4.89 \textcolor{text_red}{\scriptsize{+3.02}} 
\\

\midrule
2. $q_t \to c_t$
& WebQA        
& 18.98 & 64.59 \textcolor{text_red}{\scriptsize{+45.61}} 
& 17.46 & 61.67 \textcolor{text_red}{\scriptsize{+44.21}} 
& 21.37 & 60.34 \textcolor{text_red}{\scriptsize{+38.97}} 
& 15.42 & 49.38 \textcolor{text_red}{\scriptsize{+33.96}} 
\\

\midrule
\multirow{2}{*}{3. $q_t \to (c_i, c_t)$}
& EDIS         
& 7.13 & 31.84 \textcolor{text_red}{\scriptsize{+24.71}} 
& 6.73 & 28.96 \textcolor{text_red}{\scriptsize{+22.23}} 
& 9.54 & 25.16 \textcolor{text_red}{\scriptsize{+15.62}} 
& 5.92 & 14.71 \textcolor{text_red}{\scriptsize{+8.79}} 
\\
& WebQA        
& 2.19 & 66.43 \textcolor{text_red}{\scriptsize{+64.24}} 
& 2.05 & 68.47 \textcolor{text_red}{\scriptsize{+66.42}} 
& 4.62 & 64.95 \textcolor{text_red}{\scriptsize{+60.33}} 
& 2.02 & 59.19 \textcolor{text_red}{\scriptsize{+57.17}} 
\\

\midrule
\multirow{3}{*}{4. $q_i \to c_t$}
& VisualNews   
& 1.21 & 15.53 \textcolor{text_red}{\scriptsize{+14.32}}
& 1.18 & 15.05 \textcolor{text_red}{\scriptsize{+13.87}} 
& 2.45 & 10.88 \textcolor{text_red}{\scriptsize{+8.43}} 
& 1.05 & 6.09 \textcolor{text_red}{\scriptsize{+5.04}} 
\\
& MSCOCO       
& 15.44 & 72.62 \textcolor{text_red}{\scriptsize{+57.18}} 
& 13.86 & 66.18 \textcolor{text_red}{\scriptsize{+52.32}} 
& 14.22 & 71.16 \textcolor{text_red}{\scriptsize{+56.94}} 
& 9.18 & 44.64 \textcolor{text_red}{\scriptsize{+35.46}} 
\\
& Fashion200K  
& 1.37 & 9.70 \textcolor{text_red}{\scriptsize{+8.33}} 
& 1.41 & 12.49 \textcolor{text_red}{\scriptsize{+11.08}} 
& 0.18 & 3.93 \textcolor{text_red}{\scriptsize{+3.75}} 
& 0.15 & 4.36 \textcolor{text_red}{\scriptsize{+4.21}} 
\\

\midrule
5. $q_i \to c_i$
& NIGHTS       
& 25.80 & 28.39 \textcolor{text_red}{\scriptsize{+2.59}} 
& 25.14 & 27.71 \textcolor{text_red}{\scriptsize{+2.57}} 
& 24.67 & 30.66 \textcolor{text_red}{\scriptsize{+5.99}} 
& 19.16 & 27.09 \textcolor{text_red}{\scriptsize{+7.93}} 
\\

\midrule
\multirow{2}{*}{6. $(q_i, q_t) \to c_t$}
& OVEN         
& 0.40 & 34.26 \textcolor{text_red}{\scriptsize{+33.86}} 
& 0.35 & 33.65 \textcolor{text_red}{\scriptsize{+33.3}} 
& 0.15 & 29.77 \textcolor{text_red}{\scriptsize{+29.62}} 
& 0.02 & 25.31 \textcolor{text_red}{\scriptsize{+25.29}} 
\\
& InfoSeek     
& 0.84 & 32.87 \textcolor{text_red}{\scriptsize{+32.03}} 
& 0.79 & 34.66 \textcolor{text_red}{\scriptsize{+33.87}} 
& 0.29 & 29.23 \textcolor{text_red}{\scriptsize{+28.94}} 
& 0.18 & 16.52 \textcolor{text_red}{\scriptsize{+16.34}} 
\\

\midrule
\multirow{2}{*}{7. $(q_i, q_t) \to c_i$}
& FashionIQ    
& 1.32 & 6.35 \textcolor{text_red}{\scriptsize{+5.32}} 
& 1.03 & 6.74 \textcolor{text_red}{\scriptsize{+5.71}} 
& 1.97 & 5.96 \textcolor{text_red}{\scriptsize{+3.99}} 
& 0.89 & 4.84 \textcolor{text_red}{\scriptsize{+3.95}} 
\\
& CIRR         
& 6.09 & 16.53 \textcolor{text_red}{\scriptsize{+10.44}} 
& 4.46 & 15.94 \textcolor{text_red}{\scriptsize{+11.48}} 
& 9.06 & 15.48 \textcolor{text_red}{\scriptsize{+6.42}} 
& 7.19 & 10.36 \textcolor{text_red}{\scriptsize{+3.17}} 
\\

\midrule
\multirow{2}{*}{8. $(q_i, q_t) \to (c_i, c_t)$}
& OVEN         
& 0.16 & 48.03 \textcolor{text_red}{\scriptsize{+47.87}} 
& 0.14 & 41.65 \textcolor{text_red}{\scriptsize{+41.51}} 
& 0.93 & 40.14 \textcolor{text_red}{\scriptsize{+39.21}} 
& 0.25 & 34.57 \textcolor{text_red}{\scriptsize{+34.32}} 
\\
& InfoSeek     
& 0.14 & 43.19 \textcolor{text_red}{\scriptsize{+43.05}} 
& 0.09 & 41.11 \textcolor{text_red}{\scriptsize{+41.02}} 
& 0.22 & 38.19 \textcolor{text_red}{\scriptsize{+37.97}} 
& 0.08 & 22.94 \textcolor{text_red}{\scriptsize{+22.86}} 
\\

\midrule
- & Average   
& 6.82 & 34.69 \textcolor{text_red}{\scriptsize{+27.87}} 
& 6.28 & 33.49 \textcolor{text_red}{\scriptsize{+27.21}} 
& 7.60 & 31.83 \textcolor{text_red}{\scriptsize{+24.23}} 
& 4.67 & 23.81 \textcolor{text_red}{\scriptsize{+19.14}} 
\\
\bottomrule

\end{tabular}
}
 
    \caption{Multi-task evaluation on M-BEIR benchmark. 
    We report Recall@10 for Fashion200K and FashionIQ, and Recall@5 for others\protect \footnotemark. 
    "Base" refers to vanilla MLLMs, and "ReAlign" refers to the inclusion of our method (performance improvements are highlighted in \textcolor{text_red}{red}),
    evaluated across Qwen2-VL-7B (\textbf{Qwen2-VL-7B-Instruct}), Qwen3-VL-8B (\textbf{Qwen3-VL-8B-Instruct}), PaliGemma2-3B (\textbf{Paligemma2-3B-Mix-224}), and \textbf{Chameleon-7B}.}
    \vspace{-1mm}
    \label{tab:mbeir_result}
\end{table*}
\footnotetext{The average results of three tests are reported}

\subsection{Experiments}

\subsubsection{Experimental setup}

\paragraph{Benchmark.} We evaluate the effectiveness of our proposed \method\ on the M-BEIR benchmark~\cite{unir}. 
The benchmark comprises eight multimodal retrieval tasks involving both text and image modalities.
Based on the datasets and task configurations (the modality combinations of queries and candidates), the tasks are further categorized into 16 distinct retrieval types.
Additional details of the M-BEIR benchmark are provided in Appendix~\ref{sec:mbeir}.

\paragraph{MLLMs.} We conduct a comprehensive evaluation across four representative MLLM architectures: Qwen2-VL-7B-Instruct~\cite{qwen2vl}, Qwen3-VL-8B-Instruct~\cite{qwen3vl}, Paligemma2-3B-Mix-224~\cite{paligemma2}, and Chameleon-7B~\cite{Chameleon}.

\subsubsection{Experimental results}

\paragraph{\textit{Main Results.}}
We evaluate zero-shot multimodal retrieval on the test set of the M-BEIR benchmark. 
Table~\ref{tab:mbeir_result} demonstrates the effectiveness of our proposed \method. 
\method\ consistently yields significant performance gains across all MLLM architectures. 
The improvements are particularly profound on large-scale datasets containing massive candidate items. 
In these scenarios, vanilla MLLMs often fail to produce distinctive embeddings, resulting in inferior retrieval results.
For instance, on the InfoSeek dataset under the setting \(((q_i, c_t)\to c_i)\) (Task 6), the vanilla Qwen3-VL-8B-Instruct model achieves a negligible Recall@5 of 0.79\%. 
However, after applying \method, the performance increases to 34.66\%, marking an absolute improvement of 33.87\%. 
Similar substantial gains can be observed in all other tasks.

Additionally, we observe that retrieval performance is not strictly correlated with the model's scale or general capabilities, suggesting that a larger MLLM does not necessarily imply superior representations for retrieval. 
For example, the performance of the 7B model, Chameleon-7B, does not surpass that of the 3B model, Paligemma2-3B-Mix-224, which may be attributed to the fact that these models are optimized for dense retrieval.

\paragraph{\textit{Comparison with Stronger Baselines.}} 
To further evaluate the effectiveness of \method, we compare our method (based on Qwen2-VL-7B-Instruct) against two state-of-the-art specialized embedding models: GME~\cite{zhang2024gme} and VLM2Vec~\cite{VLM2Vec}. 
As shown in Table~\ref{tab:fine_tuned_baselines}, these models are evaluated on four representative tasks from the M-BEIR benchmark.

Despite being training-free, \method\ achieves competitive or even superior performance compared to these fine-tuned baselines. 
Notably, \method\ outperforms both GME and VLM2Vec by a significant margin in WebQA. 
For instance, \method\ reaches 66.43\% in Recall@5 on WebQA, surpassing VLM2Vec and GME by 2.53\% and 7.25\% in absolute terms, respectively. 

\begin{table}[ht]
\centering
\small
\begin{tabular}{clcccc}
\toprule
\textbf{Task} & \textbf{Dataset} & \textbf{GME} & \textbf{VLM2Vec} & \textbf{\method} \\
\midrule
1 & MSCOCO & 52.40 & 45.41 & \textbf{59.80} \\
3 & WebQA  & 59.18 & 63.90 & \textbf{66.43} \\
4 & MSCOCO & \textbf{86.70} & 39.98 & 72.62 \\
7 & OVEN  & 39.11 & \textbf{52.23} & 48.03 \\
\bottomrule
\end{tabular}
\caption{Comparison of \method\ with fine-tuned embedding baselines on four tasks from M-BEIR. All models are based on Qwen2-VL-7B architecture. We report Recall@5, and \textbf{bold} values denote the best performance.}
\label{tab:fine_tuned_baselines}
\end{table}

These results collectively validate that the representational spaces of MLLMs are a primary bottleneck for retrieval, and our proposed whitening transformation effectively aligns these representations to unlock the models' retrieval capabilities.

\section{Related works}

\subsection{Multimodal large language models}

Multimodal large language models (MLLMs) have achieved remarkable success in a wide range of tasks~\cite{mllmsurvey, zhou2024mathscape}. 
Prominent MLLMs such as LLaVA~\cite{llava,llavanext}, Qwen-VL~\cite{qwen3}, Paligemma2~\cite{paligemma2}, InternVL~\cite{internvl3_5}, and MiniCPM-V~\cite{minicpm} have shown promising advances in generative tasks such as image captioning, visual question answering, and complex multimodal reasoning~\cite{sarto2025imagecaptioningevaluationage, liang2025mathclean, guo2025brace, liu2024synthvlm, liang2024evqascore}. 

\subsection{Multimodal retrieval}

Multimodal retrieval aims to align diverse modalities (e.g., text, images, or mixed text–image) within a shared embedding space for semantic similarity matching.
Early multimodal retrieval approaches largely utilize pre-trained models such as CLIP~\cite{clip} or BLIP~\cite{BLIP} for multimodal embedding.
For instance, UniVL-DR~\cite{univldr} and UniR~\cite{unir} encode images and texts using CLIP or BLIP encoders separately, followed by fusion strategies.

With the advancement of MLLMs, researchers have begun to explore the potential of leveraging MLLMs in multimodal retrieval~\cite{LamRA, llave, unite, zhou-etal-2025-megapairs}. Specifically, a line of work focuses on building embedding models based on pretrained MLLMs to enhance multimodal retrieval, such as E5-V~\cite{E5V}, VLM2Vec~\cite{VLM2Vec}, MM-Embed~\cite{mm-embed}, and CAFe~\cite{CAFe}. 
However, these methods often require substantial computational resources and typically rely on multi-stage training strategies~\cite{Bridging, jfe}, introducing considerable costs.
In this paper, we are the first to systematically analyze the representational spaces of MLLMs from a multimodal retrieval perspective. 
Based on our observations, we propose a training-free approach that leverages intrinsic properties of MLLM embeddings to improve retrieval performance without requiring additional training.

\subsection{Whitening transformations for representations}

Whitening is a well-established technique for mitigating anisotropy in embedding spaces by de-correlating features and normalizing variance~\cite{su2021whitening, diera2024isotropy, liang2021learning}. 
In natural language processing, methods such as BERT-flow~\cite{li2020sentence} and WhiteningBERT~\cite{huang2021whiteningbert} apply whitening to transform BERT embeddings into a more isotropic distribution to improve semantic and representational properties.
In this work, we extend the application of whitening to the more intricate high-dimensional representations of MLLMs. 

\section{Conclusion}

In this work, we investigate the mechanisms that hinder MLLMs from functioning as effective zero-shot retrievers, despite their generative dominance. 
By leveraging sparse autoencoders (SAEs) to dissect the model's internal representations, we reveal that MLLMs suffer from a representational space overwhelmingly dominated by text, which suppresses essential visual information. 
Furthermore, we find that the heavy focus on bridging modalities for generation homogenizes embeddings, reducing the discriminative power required for retrieval, while the features most influential to similarity scores paradoxically act as distractors. 
Based on these insights, we propose \method, a training-free test-time adaptation method that adjusts MLLM embedding geometry. 
This simple intervention consistently and substantially improves multimodal retrieval performance across different MLLMs.

\section{Limitations}

We acknowledge some limitations in our work. 
Firstly, due to limited computational resources, we are unable to experiment with larger open-source MLLMs, such as models with 30B or 70B parameters.
Investigating how model scaling impacts the identified representational biases and the efficacy of our retrieval strategy would be a valuable direction for future research. 
Secondly, our investigation focuses primarily on the dominant image-text modality; extending the analysis and ReAlign strategy to other modalities, such as video or audio, remains a promising direction for future research.

\section*{Acknowledgments}

This work is supported by Fundamental and Interdisciplinary Disciplines Breakthrough Plan of the Ministry of Education of China (JYB2025XDXM113), National Natural Science Foundation of China (92470121, 62402016), National Key R\&D Program of China (2024YFA1014003), Zhongguancun Academy (C20250204, C20250602),  Beijing Major Science and Technology Project (Z251100008125043, Z251100008425023), and High-performance Computing Platform of Peking University.

\bibliography{custom}

\appendix

\section{Comparison of embedding extraction strategy for MLLMs}
\label{sec:embedding}

\begin{table*}[h]
    \centering
    \vspace{4pt}  
    \resizebox{0.93\textwidth}{!}{
    \begin{tabular}{lcccc|ccc|ccc}
    \toprule
    \multirow{2}{*}{\textbf{Model}} & \multirow{2}{*}{\textbf{Setting}} 
    & \multicolumn{3}{c|}{\textbf{MSCOCO}}
    & \multicolumn{3}{c|}{\textbf{NIGHTS}}
    & \multicolumn{3}{c}{\textbf{CIRR}} \\
    \cmidrule(lr){3-5} \cmidrule(lr){6-8} \cmidrule(lr){9-11} 
     &  & R@1 & R@5 & R@10 & R@1 & R@5 & R@10 & R@1 & R@5 & R@10\\ 
    \midrule
    \multirow{4}{*}{Qwen2-VL-7B-Instruct}  
    & Last token & 0.20 & 2.65 & 3.81 & 1.89 & 3.64 & 5.88 & 0.54 & 1.49 & 3.23 \\
    & Max pooling & 1.84 & 6.24 & 10.16 & 4.43 & 17.68 & 28.49 & 0.84 & 4.45 & 7.68 \\
    & Min pooling & 1.46 & 4.76 & 7.30 & 4.38 & 16.51 & 25.28 & 0.73 & 3.95 & 6.84 \\
    & Mean pooling & 5.90 & 15.44 & 22.66 & 7.55 & 22.80 & 39.11 & 1.41 & 6.09 & 9.47 \\
    \midrule
    \multirow{4}{*}{Qwen3-VL-8B-Instruct}  
    & Last token & 0.34 & 2.49 & 3.86 & 1.05 & 2.96 & 5.75 & 0.49 & 1.35 & 3.38 \\
    & Max pooling & 1.45 & 5.38 & 9.15 & 3.24 & 14.38 & 27.65 & 0.82 & 2.13 & 5.94 \\
    & Min pooling & 1.21 & 4.60 & 6.46 & 2.83 & 12.40 & 25.84 & 0.66 & 1.98 & 5.01 \\
    & Mean pooling & 4.71 & 13.86 & 21.34 & 6.46 & 20.14 & 36.42 & 1.01 & 4.46 & 7.03 \\
    \midrule
    \multirow{4}{*}{Paligemma2-3B-Mix-224}  
    & Last token & 1.63 & 3.23 & 5.81 & 1.29 & 4.26 & 8.42 & 0.43 & 1.84 & 4.16 \\
    & Max pooling & 3.26 & 7.54 & 12.37 & 5.13 & 10.84 & 17.41 & 0.81 & 4.74 & 9.50 \\
    & Min pooling & 2.75 & 5.18 & 9.84 & 3.74 & 7.10 & 13.76 & 0.64 & 3.17 & 8.02 \\
    & Mean pooling & 5.94 & 14.22 & 19.25 & 8.96 & 24.67 & 44.79 & 1.13 & 9.06 & 13.74 \\
    \bottomrule
    \end{tabular}
    }
    \caption{
    Zero-shot multimodal retrieval performance comparison between four embedding extraction strategies: \textbf{last token}, \textbf{max pooling}, \textbf{min pooling} and \textbf{mean pooling} on the MSCOCO $((q_i \to c_t))$, NIGHTS $(q_i \to c_t)$, and CIRR $((q_i, q_t) \to c_i)$ datasets.
    }
    \label{tab:mean_pooling}
\end{table*}

In this section, we explore strategies for extracting embeddings using MLLMs. We primarily focus on four strategies:
(1) \textbf{Last token}: 
This method is widely adopted in the existing literature~\cite{LamRA}. In this scenario, the multimodal input is typically constructed as \textit{"<image> <text> Summarize the above image and sentence in one word: <emb>"}, where \textit{"<image>"} and \textit{"<text>"} serve as placeholders. 
The hidden representation of the final token, denoted as \textit{"<emb>"}, is extracted as the embedding. 
While this strategy has proven effective when MLLMs are optimized for representation learning objectives, it may be suboptimal for zero-shot multimodal retrieval.
(2) \textbf{Max pooling}: 
This strategy applies max pooling to the representations from the last layer's hidden states, selecting the maximum value across the sequence dimension.
(3) \textbf{Min pooling}: 
This method constructs the embedding by selecting the minimum value across the sequence dimension of the last layer's hidden states.
(4) \textbf{Mean pooling}: 
This strategy applies mean pooling to the representations derived from the last layer's hidden states to obtain a comprehensive embedding for the entire sequence.

To evaluate these methods, we conduct comparative experiments on zero-shot multimodal retrieval. 

As shown in Table~\ref{tab:mean_pooling}, the mean pooling strategy consistently outperforms all other methods across datasets and architectures. 
This stands in sharp contrast to the last token strategy, which yields significantly inferior results.
While max and min pooling improve upon the last token baseline by aggregating information across the sequence, they still fall short of mean pooling. 
Based on these findings, we adopt mean pooling over the last layer's hidden states as the default embedding extraction method for all MLLM-based retrieval experiments in this work, unless otherwise specified.

\begin{table*}[t]
\centering
\resizebox{\textwidth}{!}{
\setlength{\tabcolsep}{2pt} 

\begin{tabular}{llllllll}
\toprule 
\textbf{Task } & \textbf{Dataset} & \textbf{Instruction (shown 1 out of 4)} & \textbf{Domain} & \textbf{Train} & \textbf{Dev} & \textbf{Test} & \textbf{Pool} \\
\midrule
\multirow{3}{*}{$1. \ q^{t} \rightarrow c^{i}$} 
& VisualNews 
& Identify news-related image match with the description & News & 99K & 20K & 20K & 542K \\
& MSCOCO 
& Find an everyday image match with caption & Misc. & 100K & 24.8K & 24.8K & 5K \\
& Fashion200K 
& Based on fashion description, retrieve matched image & Fashion & 15K & 1.7K & 1.7K & 201K \\
\midrule
$2. \ q^{t} \rightarrow c^{t}$ 
& WebQA 
& Find an paragraph from Wikipedia to answer the question & Wiki & 16K & 1.7K & 2.4K & 544K \\
\midrule
\multirow{2}{*}{$3. \ q^{t} \rightarrow (c^{i}, c^{t})$} 
& EDIS 
& Find a news image matching with the caption & News & 26K & 3.2K & 3.2K & 1M \\
& WebQA 
& Find a Wiki image that answer the question & Wiki & 17K & 1.7K & 2.5K & 403K \\
\midrule
\multirow{3}{*}{$4. \ q^{i} \rightarrow c^{t}$} 
& VisualNews 
& Provide a news-related caption for the displayed image & News & 100K & 20K & 20K & 537K \\
& MSCOCO 
& Find a caption describe the an image & Misc. & 113K & 5K & 5K & 25K \\
& Fashion200K 
& Find a description for the fashion item in the image & Fashion & 15K & 4.8K & 4.8K & 61K \\
\midrule
$5. \ q^{i} \rightarrow c^{i}$ 
& NIGHTS 
& Find an image that is identical to the given image & Misc. & 16K & 2K & 2K & 40K \\
\midrule
\multirow{2}{*}{$6. \ (q^{i}, q^{t}) \rightarrow c^{t}$} 
& OVEN 
& Retrieve a Wiki text that answer the given query about the image & Wiki & 150K & 50K & 50K & 676K \\
& InfoSeek 
& Find an article that answers the given question about the image & Wiki & 141K & 11K & 11K & 611K \\
\midrule
\multirow{2}{*}{$7. \ (q^{i}, q^{t}) \rightarrow c^{i}$} 
& FashionIQ 
& Find an image to match the fashion image and style note & Fashion & 16K & 2K & 6K & 74K \\
& CIRR 
& I'm looking for a similar everyday image with the described changes & Misc. & 26K & 2K & 4K & 21K \\
\midrule
\multirow{2}{*}{$8. \ (q^{i}, q^{t}) \rightarrow (c^{i}, c^{t})$} 
& OVEN 
& Find a Wiki image-text pair to answer a question regarding an image & Wiki & 157K & 14.7K & 14.7K & 335K \\
& InfoSeek 
& Find a Wiki image-text pair to answers my question about this image & Wiki & 143K & 17.6K & 17.6K & 481K \\
\midrule
 &\multicolumn{1}{l}{10 datasets} & \multicolumn{1}{c}{64 instructions} & \multicolumn{1}{c}{4 domains} & \multicolumn{1}{c}{1.1M} & \multicolumn{1}{c}{182K} & \multicolumn{1}{c}{190K} & \multicolumn{1}{c}{5.6M} \\
\bottomrule
\end{tabular}
}
\caption{Summary of the M-BEIR benchmarks.}
\label{tab:mbeir}
\end{table*}

\section{Training details for SAEs}
\label{appendix:sae}

In this section, we provide the comprehensive training configuration used for the sparse autoencoders (SAEs) discussed in this paper.

In experiments, we employ Top-K SAEs~\cite{topksae} to learn interpretable concept representations from the activations of the COCO~\cite{mscoco} dataset
\footnote{\url{https://huggingface.co/datasets/lmms-lab/COCO-Caption}}. 
The training process involves processing billions of activations. For instance, the training run of Qwen2-VL-7B-Instruct encompasses approximately 28 billion activations.

For MLLMs, we train SAEs on Qwen2-VL-7B-Instruct~\cite{qwen2vl}, Qwen3-VL-8B-Instruct~\cite{qwen3vl}, Paligemma2-3B-Mix-224~\cite{paligemma2} and Chameleon-7B~\cite{Chameleon}. For these architectures, SAEs are trained on the hidden states of the last layer. These SAEs differ in input dimension, but share a fixed dictionary width of 32768.
For contrastive VLMs, we utilize CLIP (ViT-Base-Patch32)~\cite{clip} and SigLIP2(SigLip2-Base-Patch16-512)~\cite{siglip2}. The SAEs for these models are trained on the derived embeddings with a dictionary width of 7168.

We train SAEs using the Adam optimizer with \(\beta_1 = 0.9\), \(\beta_2 = 0.999\), and \(\epsilon = 10^{-8}\). The learning rate is set to \(8e-4\) for all SAEs with a batch size of 4096. Our implementation utilizes the \texttt{Overcomplete}\footnote{\url{https://github.com/KempnerInstitute/overcomplete}} framework.
We conduct the training process on four NVIDIA H20 GPUs.

To mitigate the prohibitive RAM and VRAM costs associated with storing pre-computed embeddings for the entire dataset, we adopt a dynamic encoding strategy. 
Instead of pre-calculating all activations, the models encode a small proportion of the dataset on-the-fly, loading batches sequentially into GPU memory to derive the necessary activations.
To ensure training stability and prevent the optimization process from learning temporal correlations present in sequential data, we maintain a shuffled buffer of these activations, following~\cite{neel_sae_replication}. The SAEs sample training batches from this randomized buffer rather than directly from the sequential stream.

\section{Details about M-BEIR dataset}
\label{sec:mbeir}

We present the details of the M-BEIR benchmark~\cite{unir} in Table~\ref{tab:mbeir}. 
The benchmark comprises eight multimodal retrieval tasks involving both text and image modalities.
It is important to note that the M-BEIR benchmark applies additional processing to the datasets it incorporates, which may result in differences from the standard evaluation of individual datasets.
In this paper, we only utilize the test set for our evaluation.

\section{Details of evaluating learned concepts}

In Section~\ref{sec:what_hinders}, we employ sparse autoencoders (SAEs) to conduct a mechanistic analysis of MLLMs' performance in multimodal retrieval. 
While SAEs are trained on the hidden states at the token granularity to capture the fundamental feature dictionary of the model's representational space, the retrieval task itself is performed at the sample level.
In our work, the representation of a sample is typically derived via mean pooling: \(\bar{h} = \frac{1}{n} \sum_{t=1}^{n} h_t\), where \(h_t \in \mathbb{R}^d\) denotes the embedding of the \(t\)-th token.

To ensure that our analysis is directly aligned with the multimodal retrieval task, we apply SAEs to these aggregated sample-level embeddings \(\bar{h}\). Since the retrieval similarity score is calculated using the inner product of the pooled vectors in our experiments (see Appendix~\ref{sec:embedding}), any faithful attribution of this score to latent concepts must decompose the pooled vectors themselves. 
Meanwhile, although the SAE activation function is non-linear, the encoder's affine transformation is linear. Thus, the pre-activation state of a pooled embedding \(\bar{h} W_{enc}^{\top} + b\) effectively represents the arithmetic mean of token-level embeddings.
Applying SAE to individual tokens would fail to account for how features interact during the pooling process, which is critical to understanding retrieval failures.

\section{Efficiency analysis of \method}

\begin{table*}[t]
    \centering
    \small
    \setlength{\tabcolsep}{4pt} 
    \begin{tabular}{lccccc}
        \toprule
        \multirow{2}{*}{\textbf{Model}} & \multirow{2}{*}{\textbf{Hidden Dim}} & \multicolumn{2}{c}{\textbf{\method}} & \multicolumn{2}{c}{\textbf{Similarity Computation}} \\
        \cmidrule(lr){3-4} \cmidrule(lr){5-6}
        & & Time (ms) & Mem (MiB) & Time (ms) & Mem (MiB) \\
        \midrule
        PaliGemma2-3B & 2,304 & 1,538 & 43,577 & 4,107 & 43,577 \\
        Qwen2-VL-Instruct-7B & 3,584 & 3,623 & 58,978 & 5,162 & 44,677 \\
        Qwen3-VL-Instruct-7B & 4,096 & 4,533 & 65,138 & 5,369 & 48,778 \\
        \bottomrule
    \end{tabular}
    \caption{Efficiency analysis on the EDIS dataset (containing 1M candidates). We report the computation time and peak memory usage across utilizing diverse MLLMs as retrievers. The results show that \method\ introduces acceptable computation cost.}
    \label{tab:efficiency}
\end{table*}

In order to investigate the efficiency and scalability of \method, we conduct an analysis on the EDIS \((q^{t}\to(c^{i}, c^{t}))\) dataset, which contains 1 million candidate items. 

We compare the computational cost of our proposed module against the downstream similarity computation stage. 
It is important to note that a direct large-scale matrix multiplication for 1 million candidates is computationally prohibitive. 
Therefore, in our implementation, the similarity computation is optimized via batch processing. Specifically, we compute similarity matrices in small batches, extract the Top-K indices for each batch, and finally aggregate the results for evaluation.

As observed in Table~\ref{tab:efficiency}, the computational overhead of \method  \ is comparable to that of similarity search. Even as the embedding dimension increases, our framework does not exhibit significant computational bottlenecks. The results demonstrate that \method\ maintains acceptable time and memory costs, validating its efficiency and scalability for large-scale retrieval tasks.

\section{Which MLLM layer contributes the most to multimodal retrieval?}

In this section, we analyze the contribution of each layer of MLLM to multimodal retrieval performance. 
Specifically, we extract the embeddings from the hidden states of individual layers and evaluate their retrieval performance independently.

\begin{figure}[htbp]
    \centering
    \includegraphics[width=\linewidth]{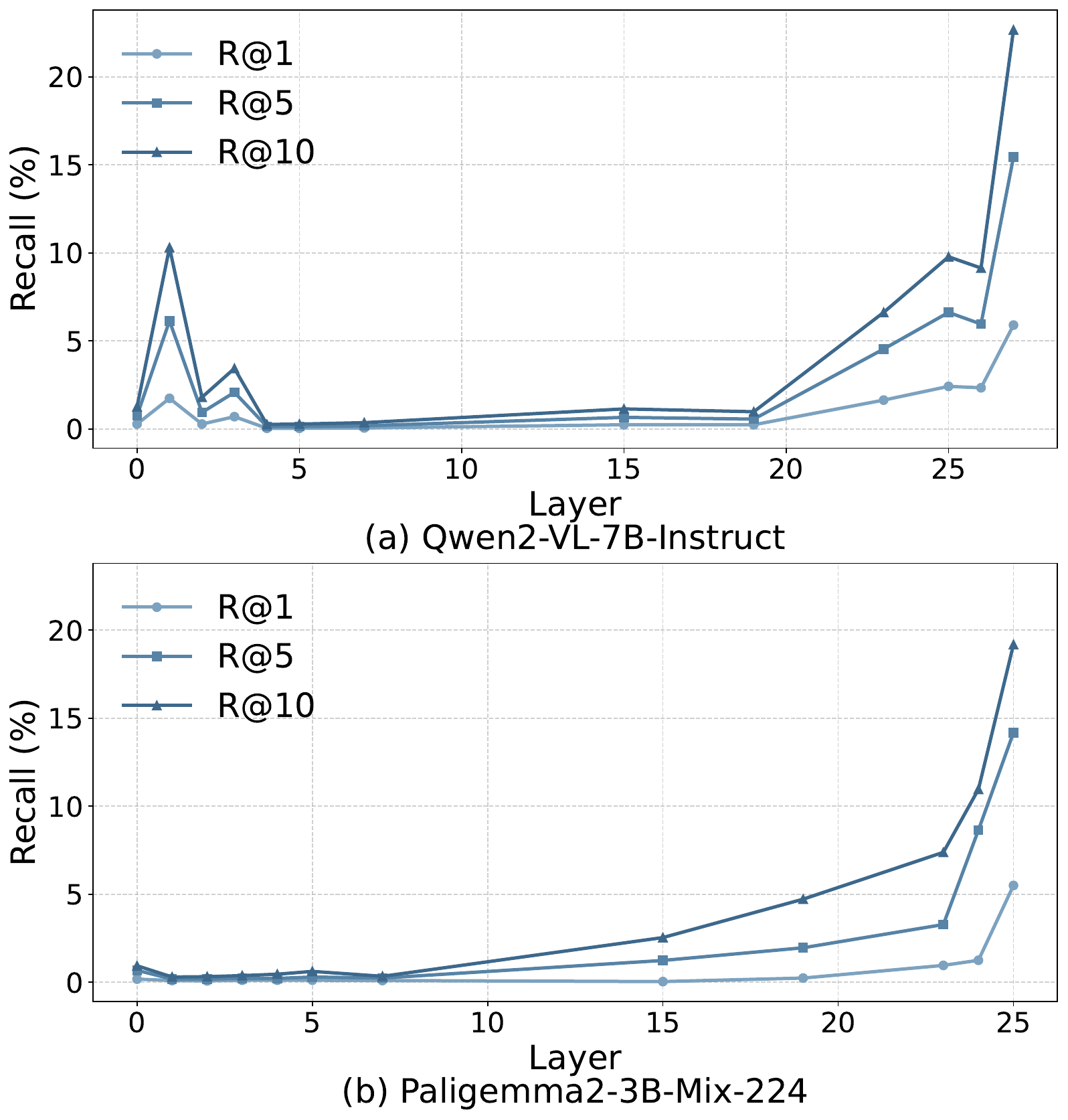}
    \caption{Retrieval performance of MLLMs across different layers on the MSCOCO $( q^{\text{i}} \rightarrow c^{\text{t}})$ dataset.}
    \label{fig:layer}
    \vspace{-1mm}
\end{figure}

As shown in Figure~\ref{fig:layer}, retrieval performance generally improves as representations are extracted from deeper layers, although progression varies between different architectures.
For Qwen2-VL-7B-Instruct, we observe a distinct performance spike in the initial layers, indicating that semantic information is captured earlier.
However, their performance remains limited compared to representations derived from deeper layers.

For both models (Qwen2-VL-7B-Instruct and Paligemma2-3B-Mix-224), embeddings from deeper layers consistently achieve higher recall across all evaluation metrics (R@1, R@5 and R@10).
In particular, the last layer representations yield the best retrieval performance.
Based on these observations, we adopt the hidden states from the final layer as the default representation for multimodal retrieval in all experiments.

\section{Parameter sensitivity analysis}

\begin{figure*}[htbp]
    \centering
    \includegraphics[width=\linewidth]{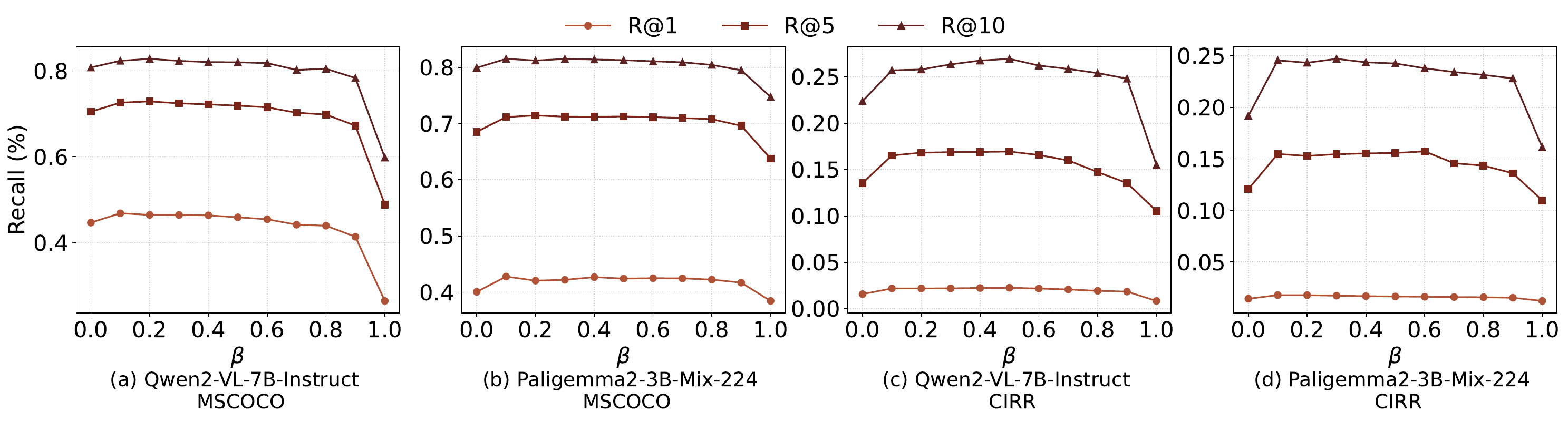}
    \caption{Impact of the shrinkage coefficient \(\beta\) on retrieval performance across the MSCOCO \((q_i \to c_t)\) and CIRR \(((q_i, q_t) \to c_i)\) datasets.}
    \label{fig:param}
    \vspace{-1mm}
\end{figure*}

We investigate how the performance of \method\ varies with respect to the parameter \(\beta\), which controls the shrinkage coefficient in the covariance estimation. 
The evaluation is conducted across the MSCOCO and CIRR datasets using two MLLM architectures under the zero-shot multimodal retrieval. The results, illustrated in Figure~\ref{fig:param}, map the retrieval performance (R@1, R@5 and R@10) as \(\beta\) varies from 0 to 1 in intervals of 0.1.

As shown in Figure~\ref{fig:param}, performance remains consistently strong within the moderate range (e.g., [0.1, 0.6]) but deteriorates at the extremes.
This indicates that while a purely empirical covariance (\(\beta=0\)) may suffer from estimation noise and a purely spherical prior (\(\beta=1\)) discards critical semantic correlations, a balanced regularization effectively stabilizes the feature geometry.

\section{Comparison of \method\ with standard PCA and ZCA in multimodal retrieval}

In this section, we analyze the effectiveness of \method\ by comparing it against the standard Principal Component Analysis (PCA) and Zero-phase Component Analysis (ZCA) baselines. 
As shown in Table~\ref{tab:zca}, \method\ consistently outperforms the standard PCA and ZCA across various MLLM backbones and datasets.

The performance gap stems primarily from the estimation strategy of covariance statistics \(\Sigma\):
The standard PCA and ZCA rely on the raw empirical covariance matrix $\Sigma$. If the embedding dimension is high (e.g., 4096), the whitening transformation can aggressively amplify noise along the directions corresponding to small eigenvalues, leading to unstable feature rectification.
Our approach mitigates these issues by employing a shrinkage estimator, interpolating between the empirical covariance and a spherical prior (identity matrix).
Empirical results demonstrate that \method\ significantly improves retrieval performance.

\begin{table*}[h]
    \centering
    \vspace{4pt}  
    \resizebox{0.93\textwidth}{!}{
    \begin{tabular}{lcccc|ccc|ccc}
    \toprule
    \multirow{2}{*}{\textbf{Model}} & \multirow{2}{*}{\textbf{Setting}} 
    & \multicolumn{3}{c|}{\textbf{MSCOCO}}
    & \multicolumn{3}{c|}{\textbf{NIGHTS}}
    & \multicolumn{3}{c}{\textbf{CIRR}} \\
    \cmidrule(lr){3-5} \cmidrule(lr){6-8} \cmidrule(lr){9-11} 
     &  & R@1 & R@5 & R@10 & R@1 & R@5 & R@10 & R@1 & R@5 & R@10\\ 
    \midrule
    \multirow{3}{*}{Qwen2-VL-7B-Instruct}  
    & PCA & 38.05 & 61.67 & 73.34 & 4.31 & 20.21 & 39.79 & 0.61 & 9.86 & 16.14 \\
    & ZCA & 44.58 & 70.50 & 80.80 & 7.75 & 26.95 & 45.53 & 1.57 & 13.57 & 22.39 \\
    & \method & 46.86 & 72.62 & 82.36 & 7.83 & 28.40 & 47.69 & 2.18 & 16.53 & 25.71 \\
    \midrule
    \multirow{3}{*}{Qwen3-VL-8B-Instruct}  
    & PCA & 30.22 & 59.13 & 73.82 & 4.11 & 22.28 & 37.36 & 0.19 & 7.24 & 14.81 \\
    & ZCA & 35.57 & 63.71 & 76.02 & 6.69 & 26.14 & 41.90 & 0.86 & 14.78 & 20.19 \\
    & \method & 38.96 & 66.18 & 77.36 & 7.83 & 27.71 & 45.83 & 1.51 & 15.94 & 24.56 \\
    \midrule
    \multirow{3}{*}{Paligemma2-3B-Mix-224} 
    & PCA & 37.43 & 62.11 & 77.30 & 5.75 & 27.46 & 49.83 & 2.01 & 19.79 & 45.13 \\
    & ZCA & 40.71 & 69.22 & 79.15 & 9.74 & 27.74 & 48.45 & 6.62 & 26.78 & 43.16 \\
    & \method & 42.76 & 71.16 & 81.52 & 9.25 & 30.66 & 49.77 & 9.26 & 30.66 & 49.77 \\
    \midrule
    \multirow{3}{*}{Chameleon-7B}  
    & PCA & 12.92 & 33.54 & 50.47 & 5.16 & 22.78 & 39.64 & 0.15 & 8.07 & 13.94 \\
    & ZCA & 19.41 & 41.56 & 56.02 & 6.98 & 24.86 & 41.03 & 0.09 & 7.76 & 14.11 \\
    & \method & 21.38 & 44.64 & 57.04 & 7.07 & 27.09 & 46.79 & 0.84 & 10.36 & 15.78 \\
    \bottomrule
    \end{tabular}
    }
    \caption{Performance comparison between the standard PCA, ZCA baselines and our proposed \method\ on the MSCOCO $((q_i \to c_t))$, NIGHTS $(q_i \to c_t)$, and CIRR $((q_i, q_t) \to c_i)$ datasets.
    }
    \label{tab:zca}
\end{table*}

\section{Comparison with contrastive VLMs}

\begin{table*}[ht!]
\centering
\small
\vspace{-1mm}
\resizebox{\textwidth}{!}{%
\begin{tabular}{llcccccccccc}
\toprule
&  & \multicolumn{2}{c}{\textbf{Contrastive VLMs}} 
& \multicolumn{2}{c}{\textbf{Qwen2-VL-7B}} 
& \multicolumn{2}{c}{\textbf{Qwen3-VL-8B}} 
& \multicolumn{2}{c}{\textbf{Paligemma2-3B}} 
& \multicolumn{2}{c}{\textbf{Chameleon-7B}} 
 \\
\cmidrule(lr){3-4}
\cmidrule(lr){5-6}
\cmidrule(lr){7-8}
\cmidrule(lr){9-10}
\cmidrule(lr){11-12}

\textbf{Task} & \textbf{Dataset}
& CLIP & SigLIP
& Base & \method
& Base & \method
& Base & \method
& Base & \method \\

\midrule
\multirow{3}{*}{1. $q_t \to c_i$}
& VisualNews   & 43.3 & 30.1 & 2.45 & 15.46 & 2.16 & 14.89 & 3.97 & 12.01 & 1.68 & 7.50 \\
& MSCOCO       & 61.1 & 75.7 & 18.57 & 59.80 & 16.39 & 57.63 & 26.63 & 66.38 & 9.74 & 48.5 \\
& Fashion200K  & 6.6  & 36.5 & 6.98 & 9.37 & 7.25 & 8.99 & 1.37 & 5.06 & 1.87 & 4.89 \\

\midrule
2. $q_t \to c_t$
& WebQA        & 36.2 & 39.8 & 18.98 & 64.59 & 17.46 & 61.67 & 21.37 & 60.34 & 15.42 & 49.38 \\

\midrule
\multirow{2}{*}{3. $q_t \to (c_i, c_t)$}
& EDIS         & 43.3 & 27.0 & 7.13 & 31.84 & 6.73 & 28.96 & 9.54 & 25.16 & 5.92 & 14.71 \\
& WebQA        & 45.1 & 43.5 & 2.19 & 66.43 & 2.05 & 68.47 & 4.62 & 64.95 & 2.02 & 59.19 \\

\midrule
\multirow{3}{*}{4. $q_i \to c_t$}
& VisualNews   & 41.3 & 30.8 & 1.21 & 15.53 & 1.18 & 15.05 & 2.45 & 10.88 & 1.09 & 6.09 \\
& MSCOCO       & 79.0 & 88.2 & 15.44 & 72.62 & 13.86 & 66.18 & 14.22 & 71.16 & 9.18 & 44.64 \\
& Fashion200K  & 7.7  & 34.2 & 1.37 & 9.70 & 1.41 & 12.49 & 0.18 & 3.93 & 0.15 & 4.36 \\

\midrule
5. $q_i \to c_t$
& NIGHTS       & 26.1 & 28.9 & 25.80 & 28.39 & 25.14 & 27.71 & 24.67 & 30.66 & 19.16 & 27.09 \\

\midrule
\multirow{2}{*}{6. $(q_i, q_t) \to c_t$}
& OVEN         & 24.2 & 29.7 & 0.40 & 34.26 & 0.35 & 33.65 & 0.15 & 29.77 & 0.02 & 25.31 \\
& InfoSeek     & 20.5 & 25.1 & 0.84 & 32.87 & 0.79 & 34.66 & 0.29 & 29.23 & 0.18 & 16.52 \\

\midrule
\multirow{2}{*}{7. $(q_i, q_t) \to c_i$}
& FashionIQ    & 7.0  & 14.4 & 1.32 & 6.35 & 1.03 & 6.74 & 1.97 & 5.96 & 0.89 & 4.84 \\
& CIRR         & 13.2 & 22.7 & 6.09 & 16.53 & 4.46 & 14.94 & 9.06 & 15.48 & 7.19 & 10.36 \\

\midrule
\multirow{2}{*}{8. $(q_i, q_t) \to (c_i, c_t)$}
& OVEN         & 38.8 & 41.7 & 0.16 & 48.03 & 0.14 & 41.65 & 0.93 & 40.14 & 0.25 & 34.57 \\
& InfoSeek     & 26.4 & 27.4 & 0.14 & 43.19 & 0.09 & 41.11 & 0.22 & 38.19 & 0.08 & 22.94 \\

\midrule
- & Average     & 32.5 & 37.2 & 6.82 & 34.69 & 6.28 & 33.49 & 7.60 & 31.83 & 4.67 & 23.81 \\
\bottomrule
\end{tabular}
}
\vspace{-1mm}
\caption{Multi-task evaluation on M-BEIR benchmark. We report Recall@5 for all datasets except Fashion200K and FashionIQ where Recall@10 is used\protect \footnotemark.}
\label{tab:vlms}
\end{table*}
\footnotetext{The average results of three tests are reported}

While our primary objective is to strengthen the multimodal retrieval capabilities of MLLMs without parameter updates, comparing \method\ with specialized contrastive VLMs (e.g., CLIP and SigLIP) offers valuable insights into the distinct advantages of different architectural paradigms. 

As shown in Table~\ref{tab:vlms}, contrastive models generally maintain a performance edge on standard image-to-text or text-to-image matching tasks (e.g., VisualNews, MSCOCO). 
This is expected, as these models are explicitly pre-trained with contrastive objectives (InfoNCE) designed to maximize the cosine similarity between text-image pairs. 
However, \method\ significantly narrows this gap, transforming MLLMs from near-random retrievers into competitive baselines. 

More importantly, our approach reveals the unique strength of MLLMs in complex and fused retrieval scenarios, especially in tasks requiring composite queries and candidates (\((q_t \to (c_i, c_t)), ((q_i, q_t) \to c_t))\)) and \(((q_i, q_t) \to (q_i, q_t))\)).
MLLMs equipped with \method\ frequently outperform contrastive baselines (e.g., Qwen2-VL with \method\ achieves 48.03 on OVEN Task 8 vs. 38.8 for CLIP). 
This suggests that \method\ successfully leverages the deep semantic comprehension capabilities inherent in MLLMs, allowing them to handle complex retrieval tasks. 
Thus, \method\ does not serve to replace specialized retrievers, but rather to equip generative giants with a robust and effective training-free retrieval mechanism.

\newpage

\end{document}